\documentclass{article}

\usepackage{amsmath,graphicx,mlspconf}
\usepackage[normalem]{ulem}
\usepackage{todonotes}
\usepackage[linesnumbered,ruled]{algorithm2e}

\usepackage[english]{babel}
\usepackage[utf8]{inputenc}
\usepackage{booktabs}
\usepackage{array}

\newcolumntype{L}[1]{>{\raggedright\arraybackslash}p{#1}}
\newcolumntype{C}[1]{>{\centering\arraybackslash}p{#1}}
\newcolumntype{R}[1]{>{\raggedleft\arraybackslash}p{#1}}

\usepackage{makecell}
\usepackage{amsfonts}
\usepackage{comment}
\usepackage{amssymb}
\usepackage{varwidth}
\usepackage[utf8]{inputenc}

\usepackage{balance}

\usepackage{times}

\copyrightnotice{978-1-5386-5477-4/18/\$31.00 {\copyright}2018 IEEE}

\toappear{2018 IEEE International Workshop on Machine Learning for Signal Processing, Sept.\ 17--20, 2018, Aalborg, Denmark}

\title{Label Propagation for Learning with Label Proportions}
\name{Rafael Poyiadzi, Raul Santos-Rodriguez and Niall Twomey\thanks{This work is supported by EurValve (Personalised Decision Support for Heart Valve Disease), Project Number: H2020 PHC-30-201 and the Continuous Behavioural Biomarkers of Cognitive Impairment (CUBOId) project funded by the UK Medical Research Council Momentum Awards under Grant MC-PC-16029.}}
\address{Intelligent Systems Laboratory, University of Bristol, BS8 1UB, United Kingdom.}

\begin{document}
\maketitle
\begin{abstract}
  Learning with Label Proportions (LLP) is the problem of recovering the underlying true labels given a dataset when the data is presented in the form of bags. This paradigm is particularly suitable in contexts where providing individual labels is expensive and label aggregates are more easily obtained. In the healthcare domain, it is a burden for a patient to keep a detailed diary of their daily routines, but often they will be amenable to provide higher level summaries of daily behavior. We present a novel and efficient graph-based algorithm that encourages local smoothness and exploits the global structure of the data, while preserving the `mass' of each bag. 
  

\end{abstract}
\section{Introduction}

The whole spectrum of learning paradigms ranging from supervised to unsupervised learning is densely packed with different settings that have limited (or no) access to the true labels. For instance, in semi-supervised learning we have access to labels only for a subset of the dataset (usually small). Additionally, we can characterize the uncertainty in the labeling process when learning from noisy and partial labels \cite{noisy,taskar}, using noise as a proxy in the former and a subset of labels per example in the later. Differently, in this paper we focus on learning from aggregated labels, also commonly referred to as Learning with Label Proportions (LLP) \cite{quadrianto2009estimating,yu2014learning}. In this setting, we assume the data comes in the form of bags of examples and for each of which we are given proportion of labels corresponding to each class. Hence, the supervision ranges from the fully supervised case (as many bags as examples) to the class priors of the whole dataset (one bag). 

Applications for LLP range from elections (labels are votes and bags are specified demographic areas) and healthcare (diagnosed diseases are released in proportions according to ZIP code area) \cite{yu2014learning}. Our motivation stems from the healthcare domain, where the Automatic Activity Recognition in Smart Home environments is key for the monitoring of health conditions such as dementia or diabetes \cite{sphere}. The task involves the classification of sensor data as belonging to a predefined set of Activities of Daily Living (ADL). The problem is usually addressed in the supervised setting, assuming that a human annotator (patient) has manually labeled enough examples of each of the classes of interest. However, this approach is not realistic when deploying this systems in the wild, as patients are reluctant to provide detailed labels. LLP offers a more amenable annotation strategy since annotations are aggregates of labels over time e.g. `today I slept 8 hours and watched TV for 2 hours'.  

\begin{figure}
  \centering
	\includegraphics[width=0.75\linewidth]{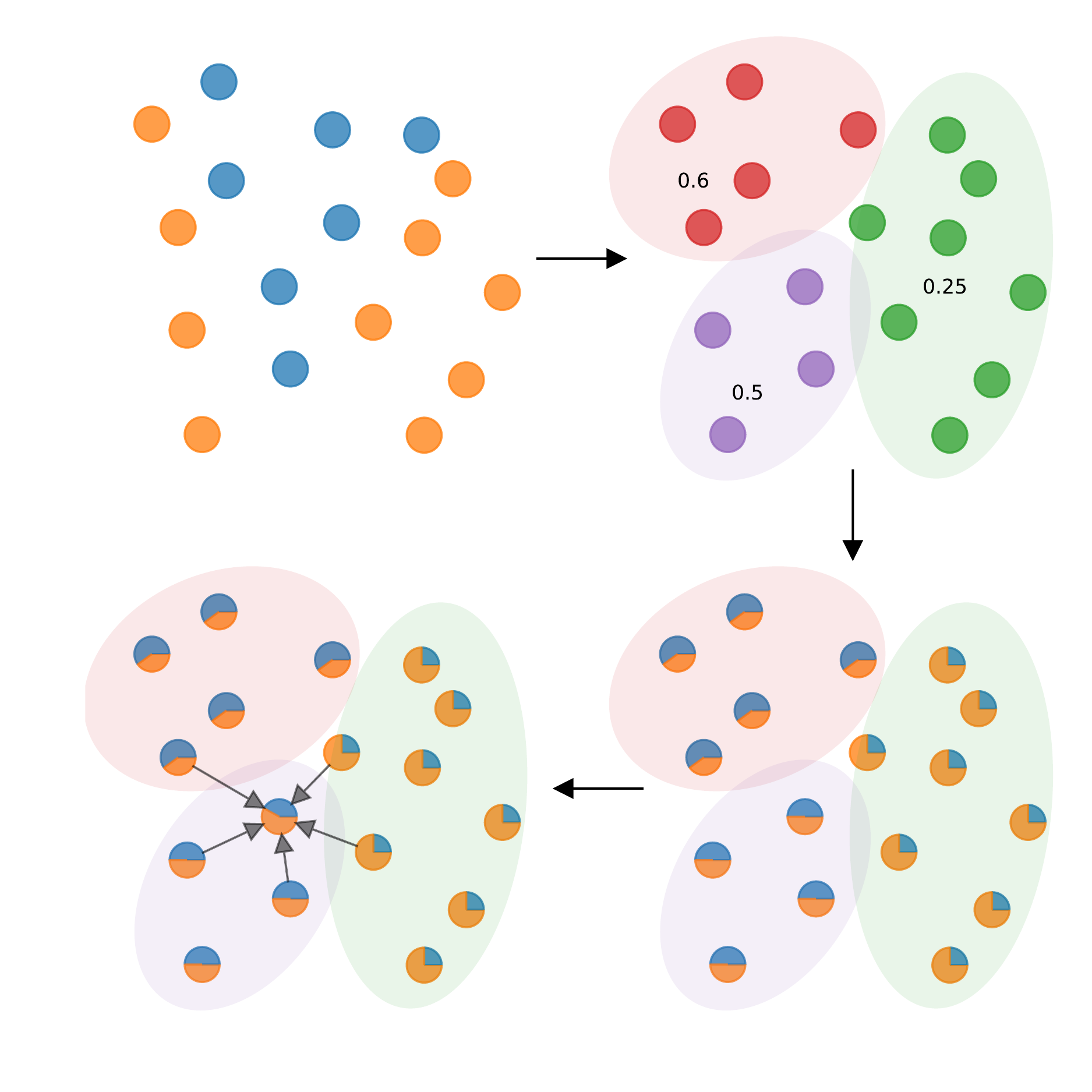}
    \caption{Illustration of the Label Propagation for Learning with Label Proportions (LP-LLP) procedure. The first step shows the bag generation and in the second step we use each bag's label proportion as a soft label for each of its points. In the last step we see that a point is influenced by its neighbors.}
	\label{fig:lp_intuition}
\end{figure}

Our proposed method is based on a graph-based formulation of the problem where data-points are represented as nodes and their relationships are encoded via edges with associated weights. Each node is initially assigned a partial label derived from its bag's bag proportion. Then each node's label is updated based on its neighbors. The problem of preserving estimated bag proportions is relaxed to the one of preserving the total `mass' of each bag. Our approach aims at recovering the true labels based on the assumption of local smoothness and by exploiting the global structure in the data.

Our contributions are twofold. Firstly, we introduce an elegant and novel graph-based approach to learning from label proportions building upon Label Propagation \cite{wang2008label}. The main intuition is illustrated in Fig. \ref{fig:lp_intuition}. Secondly, we use label proportions as the only supervision when predicting activeness in Smart Home environments for healthcare.

\textit{Problem Formulation} We assume that we have a set of observations $\textbf{X} = \{\textbf{x}_1, \cdots, \textbf{x}_n\}$ where $\textbf{x}_i \in \mathbb{R}^d$. The true labels, $\textbf{y} = \{y_1, \cdots y_n\}$ with $y_i$ being the label of observation $\textbf{x}_i$, also exist but are hidden. This set is separated into distinct bags $\textbf{X} = \bigcup_{k=1}^K \textbf{B}_k$, where each $\textbf{B}_k$ corresponds to the subset of points assigned to the $k$-th bag, and $\textbf{B}_k \cap \textbf{B}_j = \emptyset, \forall k,j \in [K]$. Moreover, for each bag $\textbf{B}_k$ we have access to its class proportions, $\boldsymbol{\pi}_k = \{\pi_{k,1}, \cdots, \pi_{k,c} \}$, where $\sum_{h=1}^c \pi_{k,h} = 1$, $\pi_{k,h} \geq 0$, with $\pi_{k,h}$ corresponding to the proportion of class $h$ in bag $\textbf{B}_k$ and $c$ being the number of classes.  

The paper is structured as follows. In Sec. 2 we review the related work. In Sec. 3 we introduce the basic formulation of Label Propagation and derive the algorithm that incorporates label proportions. In Sec. 4 we respectively present the empirical analysis in both real and synthetic datasets. Sec. 5 is devoted to the conclusions.

\section{Related Work}

Learning from bags of examples can be tracked back to tasks such as {Multiple-Instance Learning} \cite{dietterich1997solving}, where the learner is provided with logical statements indicating the presence of a class in a bag. For example, in binary classification, a bag would have a positive label if it had at least one positive point in it, while it would be labeled as negative otherwise.  

Recently, more general methods to deal with label proportions have attracted a wide interest and some of these are discussed in subsequent subsections. Existing algorithms can be loosely placed in three categories. Bayesian approaches such as \cite{kuck2012learning} approach the problem by generating labels consistent with bag proportions. In \cite{quadrianto2009estimating} the authors propose an algorithm that relies on the properties of exponential families and the convergence of the class mean operator, computed from the means and label proportions of each bag. Lastly, maximum-margin approaches \cite{yu2013propto,rueping2010svm} pose the problem as either an extension of maximum-margin clustering \cite{xu2005maximum} or Support Vector Regression. We focus on the latter two categories as they outperform the first one experimentally \cite{yu2013propto}. 
\subsection{Conditional Exponential Families}

Let $\mathcal{X}$ and $\mathcal{Y}$ denote the space of the observations and the (discrete) label space respectively, and let $\boldsymbol{\phi}(\boldsymbol{x},y) : \mathcal{X} \times \mathcal{Y} \rightarrow \mathcal{H}$ be a feature map into a Reproducing Kernel Hilbert Space $\mathcal{H}$ with kernel $k((\textbf{x},y), (\textbf{x}',y'))$. A conditional exponential family is stated as follows:
\begin{align*}
p(y|\textbf{x},\boldsymbol{\theta}) &= exp\big(\boldsymbol{\phi}(\boldsymbol{x},y)^T \boldsymbol{\theta} - g(\boldsymbol{\theta}|\boldsymbol{x})\big) \hspace{3mm} \textrm{with}\\
g(\boldsymbol{\theta}|\boldsymbol{x}) &= log \sum_{y \in \mathcal{Y}} exp\big(\boldsymbol{\phi}(\boldsymbol{x},y)^T\boldsymbol{\theta}\big)
\end{align*}
where $g(\boldsymbol{\theta}|\boldsymbol{x})$ is a log-partition function and $\boldsymbol{\theta} \in \mathbb{R}^{d}$ is the parameter of the distribution. Under the assumption that $\{(\boldsymbol{x}_i, y_i)_{i=1}^n\}$ are drawn independently and identically distributed by the distribution $p(\boldsymbol{x},y)$, one usually optimizes for $\boldsymbol{\theta}$ by minimizing the regularized negative conditional log-likelihood:
\begin{equation*}
\boldsymbol{\theta}^* = \arg \min_{\boldsymbol{\theta}} \left\{ \sum_{i=1}^n [g(\boldsymbol{\theta}|\boldsymbol{x}_i)] - n {\boldsymbol{\mu}^T_{XY}}\boldsymbol{\theta} + \lambda ||\boldsymbol{\theta}||^2 \right\},
\end{equation*}
where $\boldsymbol{\mu}_{XY} := \frac{1}{n}\sum_{i=1}^n \boldsymbol{\phi}(\boldsymbol{x}_i,y_i)$. Unfortunately, we cannot compute this quantity directly, as the labels are unknown, however in \cite{quadrianto2009estimating} the authors present \textit{MeanMap}, which makes use of the empirical means of the bags to approximate the expectations with respect to the bag distribution. This approach enjoys uniform convergence under the following (strong) assumption: conditioned on its label, a point is independent of its bag assignment $i$, namely, $p(\boldsymbol{x}|y,i) = p(\boldsymbol{x}|y)$. Extensions of MeanMap can be found in \cite{patrini2016,rafael2018a}. 

\subsection{Maximum Margin Approaches}
The maximum margin principle has been widely used in both supervised and semi-supervised learning \cite{cristianini2000introduction}. In \cite{xu2005maximum} it was also introduced to the unsupervised setting under the name Maximum Margin Clustering (MMC). 

\paragraph*{$\propto${SVM}} In \cite{yu2013propto} the authors present $\propto${SVM}, based on MMC with an extra loss function depending on the provided and estimated bag proportions. In MMC, one jointly optimizes over the labels and the separating hyperplane, with the objective of maximizing the margin between the classes, while in $\propto${SVM}, an additional constraint on the bag proportions is optimized over. The authors propose two approaches to solve the problem. The first one is of alternating nature, repeating the steps of solving the usual SVM problem and then arranging the points so that a loss on the bag proportions is minimized. Informally, the labels are arranged in such a way, such that, had an SVM been trained on the (labeled) data, it would achieve a maximum margin solution. The second approach is more convoluted and its presentation is omitted due to space.

\paragraph*{Inverse Calibration} In \cite{rueping2010svm} the authors follow the maximum margin principle by developing a model based on the Support Vector Regression. {Inverse Calibration} (InvCal) replaces the actual dataset with \textit{super-instances} \cite{yu2014learning}, one for each bag, with soft-labels corresponding to their bag-proportions.
The proportion of each bag is modeled as $q_k = (1 + exp(-\boldsymbol{w}^T\boldsymbol{m}_k + b))^{-1}$, where $\boldsymbol{m}_k = \frac{1}{|\boldsymbol{B}_k|} \sum_{\boldsymbol{x}_i \in \boldsymbol{B}_k} \boldsymbol{\phi}(\boldsymbol{x}_i)$. The constraints of the objective try to enforce $q_k \approx \pi_k$ for all bags \cite{yu2014learning}. As noted by \cite{yu2014learning} $q_k$ fails to be a good measure of bag proportions when the data has high variance, or when the distribution of the data depends on the bags, and the mean is not an adequate statistic.

\section{Label Propagation for Learning with Label Proportions (LP-LLP)}
Label Propagation is a graph-based approach to semi-supervised learning that represents the data samples as nodes on a graph and models the relationships through connecting edges with associated weights \cite{wang2008label}. In this section we will first present LP in the semisupervised setting following \cite{wang2008label,zhou2004learning} and then adopt the procedure to LLP. We then provide a comparison with the $k$-Nearest Neighbours approach.

\subsection{Label Propagation}
 Let $\boldsymbol{F}, \boldsymbol{Y} \in \mathbb{R}^{n\times c}$, where $c$ is the number of classes. $\boldsymbol{F}$ corresponds to a `soft' label matrix with each row corresponding to one point, and each element of that row corresponds to the probability of assigning the particular data sample to its class. $\boldsymbol{Y}$ encodes the true labels with $\boldsymbol{Y}_{ij} = 1$ if $y_i = j$ and 0 otherwise. Our method is presented in Algorithm~\ref{generalLP}.

\begin{algorithm}
    \caption{Label Propagation}
	\label{generalLP}
    \SetKwInOut{Input}{Input}
    \SetKwInOut{Output}{Output}

Compute a similarity matrix $\boldsymbol{W}$ for the whole dataset. Restrict the diagonal to $W_{ii} = 0$.
    
Let $\boldsymbol{D}$ be a diagonal matrix with $\boldsymbol{D}_{ii}$ being the sum of the $i$-th row of $\boldsymbol{W}$. Compute $\boldsymbol{S} = \boldsymbol{D}^{-1}\boldsymbol{W}$. 
    
Let $0 < \alpha < 1$ and iterate $F(t+1) = \alpha \boldsymbol{S}F(t) + (1 - \alpha)\boldsymbol{Y}$. $(\boldsymbol{F}(0) = \boldsymbol{Y})$
   
Let $F^*$ denote the limit of $F(t)$. Assign labels based on $y_i = \arg \max_{j} F_{ij}^*$. 
\end{algorithm}

The first step assigns weights to the edges based on a similarity function. A popular choice is $\boldsymbol{W}_{ij} = exp(-\gamma ||\boldsymbol{x}_i - \boldsymbol{x}_j||^2)$, with $\gamma > 0$. The second step normalizes the similarity matrix such that each row sums to 1. In the third step $\alpha$ can be thought of as controlling how the information coming from your neighbors and the information coming from the labeled examples are weighted. In the final step labels are assigned.

It can be shown \cite{zhou2004learning} that the sequence convergences to $\boldsymbol{F}^* = (1 - \alpha)(\boldsymbol{I} - \alpha \boldsymbol{S})^{-1}Y$.

In the following, without loss of generality, we restrict the problem to the binary case for simplicity. We will now use $\boldsymbol{f} \in [-1,1]^{n}$ to denote the predictions. An individual update is as follows:
\begin{equation}
\label{single_update}
f_i^{t+1} = \alpha \sum_{j} \bar{w}_{ij} f_j^t + (1 - \alpha)y_i,
\end{equation}
where $\bar{w}_{ij}=w_{ij}/\sum_{j} w_{ij}$. 

We now provide a different interpretation through the analogy of a random walk. We have $\boldsymbol{f}^* = (\boldsymbol{I} - \alpha\boldsymbol{S})^{-1}\boldsymbol{y}$, which as we have already seen, can be formulated as $\boldsymbol{f}^* = \sum_{i=1}^{\infty} (\alpha\boldsymbol{S})^{i}\boldsymbol{y}$. Think of $\boldsymbol{S}$ as a stochastic matrix and consider a walker sitting on the $i$-th node about to decide where to move next based on the $i$-th row of $\boldsymbol{S}$. What is the probability that on the \textit{next step} the walker lands on a positive node? These (unnormalized) probabilities can be computed by raising $\boldsymbol{S}$ to the corresponding power. Imagine keeping track of the count of positive and negative visits, and using them to provide an estimate for the label of the $i$-th node. But of course, earlier visits are more important as they have a higher probability of being close to $i$. This weighting is given by the value of $\alpha$.

\subsection{Label Propagation with Label Proportions}
In the label proportions setting \textit{no label} is provided for any of the points. Using the walker analogy, after the first step, the walker has landed on node $j$, where $\boldsymbol{x}_j \in \textbf{B}_k$ (to avoid confusion, we imply that sample $\boldsymbol{x}_j$ is assigned to the $k$-th bag. However, in this case, we do not have access to the true label of $\boldsymbol{x}_j$). What we exploit in this paper is the idea that even if we do not actually know the true label of $j$, we have access to its bag's label proportions, which we cast as prior probability of being assigned to a class. The intuition of a random walk in semisupervised learning carries on, only now, we have replaced the `hard' labels with `soft' labels, that loosely represent probabilities. 
In the binary classification case, our $\boldsymbol{\hat{y}}$ can now be defined as $\hat{y}_i = \pi_{k,1}$ where $x_i \in \textbf{B}_k$ and $\pi_{k,1}$ represents the proportion of positive labels ($1$) in bag $k$ ($\pi_{k,0} = 1 - \pi_{k,1}$ is equivalently the proportion of negative labels ($0$) in the same bag). One could now compute $\boldsymbol{f}^* = (\boldsymbol{I} - \alpha\boldsymbol{S})^{-1}\boldsymbol{\hat{y}}$, but trivial decision making on $\hat{y}$ does not guarantee preservation of the class proportions.

Let us now have a look at the problem from a regularization perspective \cite{wang2008label} as this will allow us to introduce the bag proportions as constraints in a principled manner. Consider the following loss function that we want to minimize.
\begin{equation*}
Q(\boldsymbol{f}) = \sum_{i=1}^n \sum_{j=1}^{n} s_{ij} (f_i - f_j)^2 + \gamma \sum_{i=1}^n (f_i - y_i)^2 
\end{equation*}
where the first term encourages local smoothness while the second penalizes deviation from $y$. The balance between the two is controlled through $\gamma$. The solution of $\arg \min_f Q(f)$ gives the same solution as before \cite{wang2008label}. In its original form, the problem would be an Integer Program, $\arg \min_{f \in \{0,1\}^n} Q(f)$, which is in general intractable. Building on this, one could enforce bag constraints through a system of linear equations $\boldsymbol{A}\boldsymbol{f} = \boldsymbol{b}$, where $\boldsymbol{A} \in \mathbb{R}^{K \times n}$, with $K$ being the number of bags and $\boldsymbol{b} \in \mathbb{R}^{K}$. $\boldsymbol{A}$ is defined as $\boldsymbol{A}_{ji} = 1$, if $x_i \in \boldsymbol{B}_j$ and 0 otherwise, and $\boldsymbol{b}_j = \boldsymbol{n}_{j,1}$, where $\boldsymbol{n}_{k,c}$ corresponds to the number of instances of class $c$ in  $\boldsymbol{B}_k$. 

We now have a constrained Integer Program and proceed by relaxing it to a constrained Linear Program as follows. Instead of controlling the exact number of points for each class for each bag, we are instead controlling the total `mass' of each class in each bag. With $\boldsymbol{f}$ unconstrained, any $\boldsymbol{f}_i$ can dominate over the others, rendering the `mass' conservation principle useless. A more suitable constraint would then be $\boldsymbol{f} \in [0,1]^n$ instead, giving our final problem formulation:
\begin{align}
\boldsymbol{f}^* &= \arg \min_{\boldsymbol{f} \in [0,1]^n} \boldsymbol{Q}(f) \nonumber\\
s.t.\mbox{ } & \boldsymbol{A}\boldsymbol{f} = \boldsymbol{b}
\label{eq:optimization}
\end{align}

\paragraph*{Label Propagation for Learning with Label Proportions} The proposed algorithm solves Eq.~\ref{eq:optimization} in two steps, by first solving the {unconstrained} problem $\boldsymbol{f} = (\boldsymbol{I} - \alpha\boldsymbol{S})^{-1}\boldsymbol{\hat{y}}$ (with $\hat{y}_i = \pi_{k,1}$ where $x_i \in \textbf{B}_k$), and then applying \textit{Alternating Projections} \cite{boyd2003alternating} to determine $\boldsymbol{f}^*$ such that $\boldsymbol{A}\boldsymbol{f}^* = \boldsymbol{b}$. 
Alternating Projections is a simple approach for finding a point in the intersection of convex sets. In our case the sets correspond to row and column sums, as each row of $\boldsymbol{F}$ should live on a probability simplex and column comes from the bag proportions, respectively. We refer the reader to \cite{boyd2003alternating} and the references therein for convergence guarantees.

These two steps are repeated until convergence, with $\boldsymbol{f}^{(t+1)} = (\boldsymbol{I} - \alpha\boldsymbol{S})^{-1}\boldsymbol{f}^{(t)}$, and $\boldsymbol{f}^{(0)} = \boldsymbol{\hat{y}}$. 

The procedure for solving the optimization problem in Eq.~\ref{eq:optimization} is depicted in Alg.~\ref{algo}. 
\begin{algorithm}
    \caption{Label Propagation for Learning with Label Proportions (binary case)}
	\label{algo}
    \SetKwInOut{Input}{Input}
    \SetKwInOut{Output}{Output}

    \Input{Bag assignment matrix $\boldsymbol{A}$ and vector $\boldsymbol{b}$ with $\boldsymbol{b}_j = \boldsymbol{n}_{j,1}$}
    \Output{Estimated labels $\boldsymbol{\hat{f}}$}
    Compute similarity matrix $\boldsymbol{W}$ and then $\boldsymbol{S}$ \\
    Compute $\boldsymbol{f}^{(t+1)} = (\boldsymbol{I} - \alpha\boldsymbol{S})^{-1}\boldsymbol{f}^{(t)}$, where $\boldsymbol{f}^{(0)}$ is defined as: $f^{(0)}_i = \pi_{k,1}$ for $\boldsymbol{x}_i \in \boldsymbol{B}_k$, $\forall i$.\\
    Solve for $f^{{(t+1)}^*}$ using Alternating Projections. \\
    Repeat steps (2) and (3) until convergence. \\
    Estimate labels based on $\hat{f}_i = sgn(f^*_i - 0.5)$
\end{algorithm}

\paragraph*{Weighted k-Nearest Neighbors (k-NN)} We now analyze the connection between LP-LLP and the weighted versions of k-NN. Referring back to the LP formulation we showed that $$\boldsymbol{f}^* = (\boldsymbol{I} - \alpha\boldsymbol{S})^{-1}\boldsymbol{y} = \sum_{k=0}^{\infty} (\alpha\boldsymbol{S})^{k}\boldsymbol{y} = \boldsymbol{y} +  (\alpha\boldsymbol{S})\boldsymbol{y} + (\alpha\boldsymbol{S})^{2}\boldsymbol{y} + \cdots$$ For one sample, and considering already normalized weights to ease the notation, we have:
\begin{equation}
f_i = y_i + \alpha \sum_{j \in \mathcal{N}(i)} \bar{w}_{ij} y_j + {\alpha}^2 \sum_{k \in \mathcal{N}(i) \cap \mathcal{N}(j)} \bar{w}_{ik} \bar{w}_{kj} y_j + \cdots\nonumber
\end{equation}

$\mathcal{N}(i)$ refers to the neighborhood of $i$, that is $w_{ij} \neq 0$. 
\begin{equation}f_i = \sum_{j \in \mathcal{N}(i)} y_j \left[\alpha \bar{w}_{ij} + {\alpha}^2 \sum_{h \in \mathcal{N}(i) \cap \mathcal{N}(j)} \bar{w}_{ih} \bar{w}_{hj} + \cdots \right]\nonumber
\end{equation}
In a weighted k-NN setting, the updates would be of the form $\dot{f}_i = \sum_{j \in \mathcal{N}(i)} \bar{w}_{ij} y_j$. Ignoring $\alpha$ for now, we see that the first term of the LP update is equivalent to the k-NN update. This term encourages nearby points to have the same label (\textit{local smoothness}). Looking at the second term of the LP updates we see that it is an iteration over common neighbors of nodes $i$ and $j$ (node $i$ is the node we wish to update, while node $j$ is one of the neighbors we will be `influenced' by). 
If nodes $i$ and $j$ have close common neighbors, then this will enhance the `influence' of $j$ on $i$. We can understand this as an exploitation of the \textit{global structure} of our data, points lying on the same structure, should have the same label. These two attributes, local smoothness and global structure, are the underlying principles of many semisupervised learning algorithms, and LP-LLP brings them to the Learning with Label Proportions setting.

\paragraph*{Time Complexity} The algorithm requires the computation of a similarity matrix which would require $\mathcal{O}(N^2)$, where $N$ is the number of data points, and then compute the generalized Laplacian. The bottleneck is computing its inverse which has complexity $\mathcal{O}(N^3)$. Similarly to other well-established machine learning algorithms which share this bottleneck, one could make use of approximations that would trade off accuracy for computational expenses~\cite{kishore2017literature}. We also note that the per iteration complexity scales linearly in $N$, due to the normalization step.

\subsection{Comparison of Methods}
A critical assumption of MeanMap is the conditional independence of a data sample and its bag assignment, given its label. Moreover, the practitioner has no control over the estimated bag proportions. Some of the merits of this approach include probabilistic labels, principled handling of multiclass classification problems and a straightforward implementation. Moreover, the techniques that rely on the max-margin principle lose probabilistic labels but these are not limited to the dependence of data generation to bags. Unlike, $\propto$-SVM, InvCal cannot handle multiclass problems naturally and does not offer a direct control over bag proportions. It is, however, easy and cheap to implement, whereas $\propto$-SVM is more convoluted.

Our proposed graph-based method does not make assumptions on the bag generation process and can seamlessly handle multiclass classification tasks. It does not offer probabilistic labels, but instead it provides constrained values that can be interpreted as ranks. These are then used to provide a degree of control over the estimated bag-proportions.

\section{Experiments}
In our experiments we consider synthetic and real-world datasets. We first describe both and then present the results.

\begin{figure}
  \centering
	\includegraphics[width=1\columnwidth]{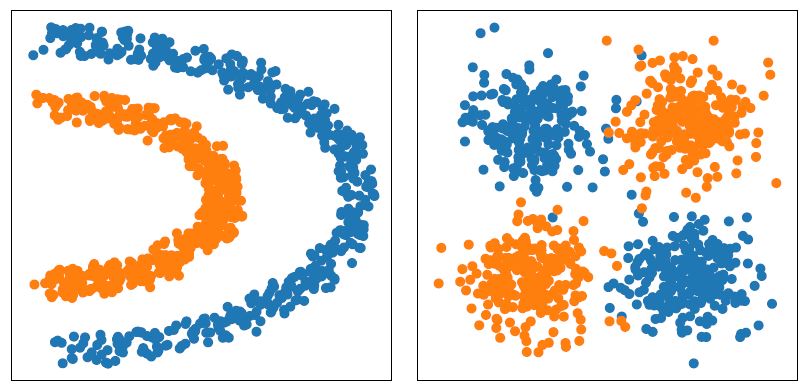}
    \caption{The two synthetic datasets considered in this paper. From left to right, Half Kernel and XOR.}
	\label{fig:synthetic}
\end{figure}

\paragraph*{Synthetic Datasets} We have two different synthetic datasets shown in Fig.~\ref{fig:synthetic}. The XOR dataset was created by generating four Gaussian distributions with means lying on the corners of a square with size size of 10 and identity covariance matrix. The Half-Kernel dataset was generated by a weighted combination of trigonometric functions, \textit{sin} and \textit{cos}, and added noise.

\paragraph*{Real-World Dataset} In the Sphere Challenge dataset \cite{twomey2016sphere}, sensor data (including accelerometer, video, and environmental) was collected from 10 people on two separate occasions. 
There were 8 males and 2 females, with 8 between the ages of 18 to 29 and 2 within the ages of 30 to 39. Each participant was wearing a wrist-worn accelerometer and was asked to complete a series of scripted activities, taking around 25 to 30 minutes in total. The participants performed several activities included that are categorized into ambulation activities (e.g. walking), posture activities (e.g. standing) and transitional activities (e.g. sit to stand).
This script was carried out twice in full by each participant on different days, and each was annotated at least twice, producing labels which are themselves proportions. 
Our experiments consider the task of classifying between ambulatory and sedentary activities. 
We consider two settings of this dataset. Setting 1 involves predicting ADLs from the raw sensor data, while Setting 2 involves predicting ADLs from the simple baseline features of the dataset, as described in \cite{twomey2016sphere}. 

\paragraph*{Model Setup} We focus only on binary classification problems in this work since this is the setting of our baseline methods. 
The data samples into three bags. In the case of synthetic data, the data is first generated according to the desired total size and then separated into the bags, respecting the desired bag proportions. 
We consider that test data is provided in the format of a bag with known bag proportion. 
The approach we take is to train our models \textit{with} the test data included, ignoring the true label of each instance.
For the experiments we fix parameter $\alpha$ to $0.50$ and choose $\gamma$ by running the algorithm on a grid of values and then choosing based on the heuristic of highest score in terms of $\boldsymbol{\bar{f}}^T\boldsymbol{S}\boldsymbol{\bar{f}}$.  
We compare LP-LLP against $\propto$SVM and InvCal. MeanMap is not included experiments mainly because its application is not universal, due to its assumptions, and secondly because it has been shown to perform slightly worse than other approaches\cite{yu2013propto,rueping2010svm}. Each data-point corresponds to features extracted (such as, mean, min, max etc) from accelerometer data from a small time-window. The data-points were annotated so true labels exist. We ignore the time aspect of the data and separate the data-points in bags at random such that each bag has a pre-specified proportion of each class.

\paragraph*{Discussion} 
Tables \ref{table:xor}, \ref{table:hk}, \ref{table:sphere1} and \ref{table:sphere2} present our experimental results. The number of experiments performed per dataset was 25.
These should be read as follows. The first column describes the format of the dataset, for example, $'100A'$ corresponds to a dataset consisting of a training set of size $100$ and a test set of size $20\%$ of that. The letter denotes the bag proportion configuration (recall we have separated our data into 3 bags), $A$ - $(0.60, 0.40, 0.50)$ and $B$ - $(0.85, 0.25, 0.40)$. The rest of the values should be read as \textit{mean}(\textit{one standard deviation}).
Finally, the model that performs best is emboldened in the results tables. (We would like to point out that for Table 3, $\propto$-SVM did not converge.) As seen empirically our proposed method, in general, compares favorably with alternative approaches over the broad set of experimental configurations considered. With regards to synthetic data, we see that even if $\propto$SVM has a low predictive mean accuracy, it has a high standard deviation. This is due to the method confusing the two classes. In general the performance of all three methods improves with increasingly large datasets, as expected. We note that the Sphere2 configuration of the dataset has the most varied performance (Table \ref{table:sphere2}). We believe that this is a more challenging representation for all models (due the its higher dimensionality) and this is illustrated by higher predictive variance on all models.

\begin{table}[!h]
    \begin{minipage}{\columnwidth}
      \caption{XOR}
        \label{table:xor}
      \centering
          \begin{tabular}{lccc}
           \toprule
            & $\propto$SVM & InvCal & LP-LLP\\
           \midrule

           120A & 0.54(0.19) & 0.81(0.08) & \textbf{0.91(0.14)}\\
           120B & 0.66(0.46) & 0.73(0.05) & \textbf{0.97(0.04)}\\

           180A & 0.50(0.04) & 0.83(0.08) & \textbf{0.88(0.18)} \\
           180B & 0.42(0.09) & 0.73(0.06) & \textbf{0.97(0.03)}\\

           300A & 0.49(0.06) & 0.88(0.13) & \textbf{0.92(0.12)}\\
           300B & 0.56(0.07) & 0.73(0.04) & \textbf{1.00(0.01)}\\

           600A & 0.45(0.43) & 0.94(0.04) & \textbf{0.99(0.01)}\\
           600B & 0.67(0.46) & 0.69(0.04) & \textbf{0.99(0.01)}\\
           \bottomrule
          \end{tabular}
          \vspace{0.1em}
\end{minipage}%

\begin{minipage}{\columnwidth}
      \centering
        \caption{Half-Kernel}
        \label{table:hk}
          \begin{tabular}{lccc}
         \toprule
          &$\propto$SVM & InvCal & LP-LLP \\
         \midrule
         120A & \textbf{0.81(0.08)} & 0.46(0.26) & 0.58(0.16)\\
         120B & 0.73(0.05) & 0.56(0.31) & \textbf{0.74(0.18)}\\

		 180A & \textbf{0.83(0.08)} & 0.63(0.06) & 0.59(0.13)\\
         180B & 0.73(0.06) & 0.50(0.50) & \textbf{0.93(0.13)}\\

         300A & \textbf{0.88(0.08)} & 0.67(0.08) & 0.84(0.19)\\
         300B & 0.73(0.04) & 0.50(0.50) & \textbf{0.96(0.08)}\\

         600A & 0.94(0.04) & 0.70(0.08) & \textbf{0.96(0.05)}\\
         600B & 0.69(0.04) & 0.74(0.05) & \textbf{1.00(0.00)}\\
         \bottomrule
         \end{tabular}
          \vspace{0.1em}
    \end{minipage}

    \begin{minipage}{\columnwidth}
      \caption{Sphere 1}
        \label{table:sphere1}
      \centering
          \begin{tabular}{lccc}
           \toprule
            & $\propto$SVM & InvCal & LP-LLP\\
			\midrule
           60A & 0.50(0.00) & 0.58(0.29) & \textbf{0.76(0.39)}\\
           60B & 0.50(0.00) & 0.50(0.00) & \textbf{0.88(0.29)}\\ 

           120A & 0.50(0.00) & 0.54(0.28) & \textbf{0.59(0.45)} \\
           120B & 0.50(0.00) & 0.50(0.00) & \textbf{1.00(0.00)}\\

           180A & 0.52(0.05) & 0.56(0.26) & \textbf{0.80(0.26)} \\
           180B & 0.50(0.00) & 0.50(0.00) & \textbf{1.00(0.01)} \\

           240A & 0.50(0.00) & 0.62(0.19) & \textbf{0.78(0.16)} \\
           240B & 0.50(0.00) & 0.50(0.00) & \textbf{0.98(0.01)}\\
           \bottomrule
          \end{tabular}

          \vspace{0.1em}
    \end{minipage}%

    \begin{minipage}{\columnwidth}
      \centering
        \caption{Sphere 2}
        \label{table:sphere2}
         \begin{tabular}{lccc}
         \toprule
          &$\propto$SVM & InvCal & LP-LLP\\
         \midrule
         60A & 0.48(0.03) & \textbf{0.54(0.09)} & 0.53(0.09) \\
         60B & 0.47(0.04) & \textbf{0.54(0.07)} & 0.50(0.09) \\

         120A & 0.47(0.11) & 0.47(0.12) & \textbf{0.50(0.13)} \\
         120B & 0.44(0.09) & 0.57(0.10) & \textbf{0.58(0.12)} \\

         180A & 0.60(0.11) & \textbf{0.65(0.14)} & 0.60(0.16) \\
         180B & 0.51(0.10) & 0.57(0.13) & \textbf{0.65(0.16)} \\

         240A & \textbf{0.68(0.16)} & 0.58(0.19) & 0.56(0.20) \\
         240B & 0.34(0.19) & 0.54(0.10) & \textbf{0.74(0.12)} \\
         \bottomrule
         \end{tabular}
    \end{minipage} 
\vspace{-4mm}
\end{table}

\section{Conclusion}
We have introduced a new approach to the problem of learning with label proportions based on a graph-based representation of the data. Our approach encourages local smoothness and exploits global structure to assign labels. This is justified both qualitatively and empirically. Experiments carried out demonstrate the performance of our approach as compared to existing algorithms. Our method is not constrained by the generation of the bags, as opposed to MeanMap. By taking a transductive approach to training we tend to overfit less, which is not always the case for $\propto$SVM and InvCal. 


\bibliographystyle{IEEEbib}
\bibliography{bibliography}
\end{document}